%% file: root.tex
\documentclass[letterpaper,10pt,conference]{ieeeconf}

\IEEEoverridecommandlockouts
\usepackage[utf8]{inputenc}
\usepackage[T1]{fontenc}
\DeclareUnicodeCharacter{1D70B}{\ensuremath{\pi}}
\DeclareUnicodeCharacter{03C0}{\ensuremath{\pi}}
\DeclareUnicodeCharacter{2080}{\textsubscript{0}}
\DeclareUnicodeCharacter{2019}{'}
\DeclareUnicodeCharacter{2014}{--}
\DeclareUnicodeCharacter{2217}{*}
\usepackage{amsfonts}
\usepackage{amsmath}
\usepackage{amssymb}
\usepackage{booktabs}
\usepackage{capt-of}
\usepackage{cuted}
\makeatletter
\@ifundefined{@makespecialcolbox}
  {\let\@makespecialcolbox\@make@specialcolbox}
  {}
\makeatother
\usepackage{graphicx}
\usepackage{microtype}
\usepackage{multirow}
\usepackage{nicefrac}
\usepackage{placeins}
\usepackage{url}
\usepackage{xcolor}

\input{commands}

\title{\LARGE \bf
Modality-Autoregressive World-Action Models
}

\author{Adam Hung, Bardienus P. Duisterhof, Deva Ramanan, Jeffrey Ichnowski\\
Carnegie Mellon University\\
\textbf{Project page: \textcolor{modarred}{\url{https://adamhung60.github.io/ModAR/}}}}

\begin{document}
\bstctlcite{IEEEexample:BSTcontrol}

\maketitle
\thispagestyle{empty}
\pagestyle{empty}

\setlength{\stripsep}{0pt}
\begin{strip}
    \vspace{-20pt}
    \centering
    \includegraphics[width=\textwidth]{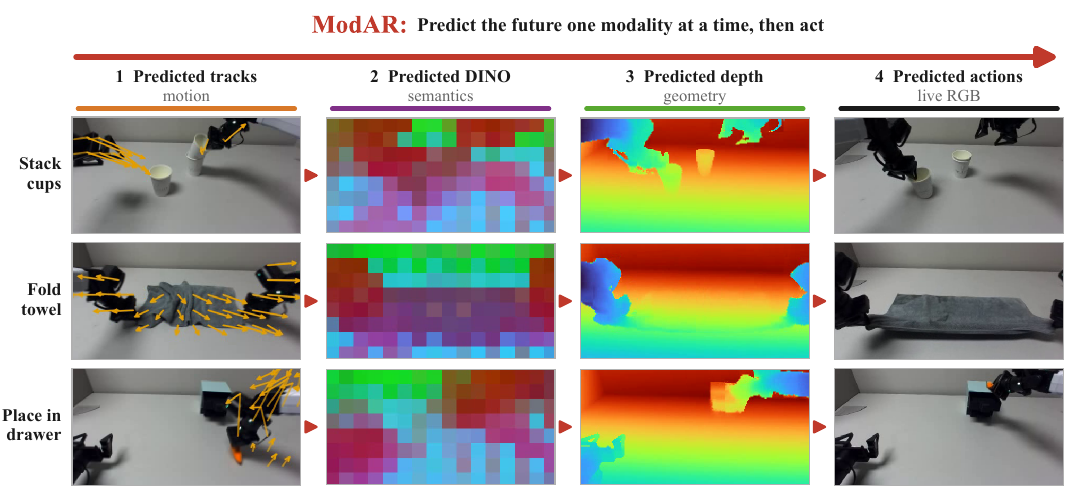}
    \captionof{figure}{We present \methodname, a world-action model (WAM) that predicts future observations as a sequence of multiple modalities. 
    \methodname{} autoregressively denoises one modality at a time before finally denoising robot actions.
    We train all models from scratch for controlled comparisons across predicted modalities and training data.
    We find that additional modalities improve performance, and our multimodal formulation outperforms prior WAM formulations. We further validate \methodname{} on real-world bimanual manipulation using robot demonstrations and human videos.
    \vspace{8pt}
    }

    \label{fig:teaser}
\end{strip}

\input{0_abstract}
\input{1_intro}
\input{2_rw}
\input{3_method}

\input{4_experiments}
\input{5_conclusion}

\FloatBarrier
\bibliographystyle{IEEEtran}
\bibliography{references}

\end{document}

%% file: commands.tex
\newcommand{\methodname}{ModAR}
\newcommand{\best}[1]{\textbf{#1}}
\newcommand{\second}[1]{\underline{#1}}
\definecolor{modarred}{HTML}{CC3829}

%% file: 0_abstract.tex
\begin{abstract}
World-action models (WAMs) jointly model future observations and actions, typically predicting the future as RGB images.
Other visual modalities such as depth, pretrained visual features, and point tracks can more efficiently capture geometric, semantic, and motion features.
However, how best to combine these modalities within WAMs remains an open question.
We introduce \methodname{}, the first WAM to autoregressively denoise multiple future modalities before predicting actions. 
This allows each prediction to condition on previously generated modalities. 
We train from scratch to systematically study how training-data mixtures, predicted modalities, and WAM formulations affect performance.
In our evaluations, WAMs benefit from predicting point tracks, DINO features, and depth maps, while additionally predicting future RGB does not provide a consistent benefit.
We also find that \methodname{}'s sequential generation outperforms existing WAM formulations, with the highest average success rate at all evaluated data scales.
We also fine-tune the video-model-initialized WAM Flex-$\pi$ on the same data; \methodname\ achieves a slightly higher observed average success rate (75\% vs.\ 72\%) while using approximately \(20\times\) fewer training FLOPs and no pretraining.
On three real-world bimanual tasks, \methodname\ outperforms baselines and improves with human videos.

\end{abstract}

%% file: 1_intro.tex
\begin{figure*}[t]
    \centering
    \includegraphics[width=0.8\linewidth]{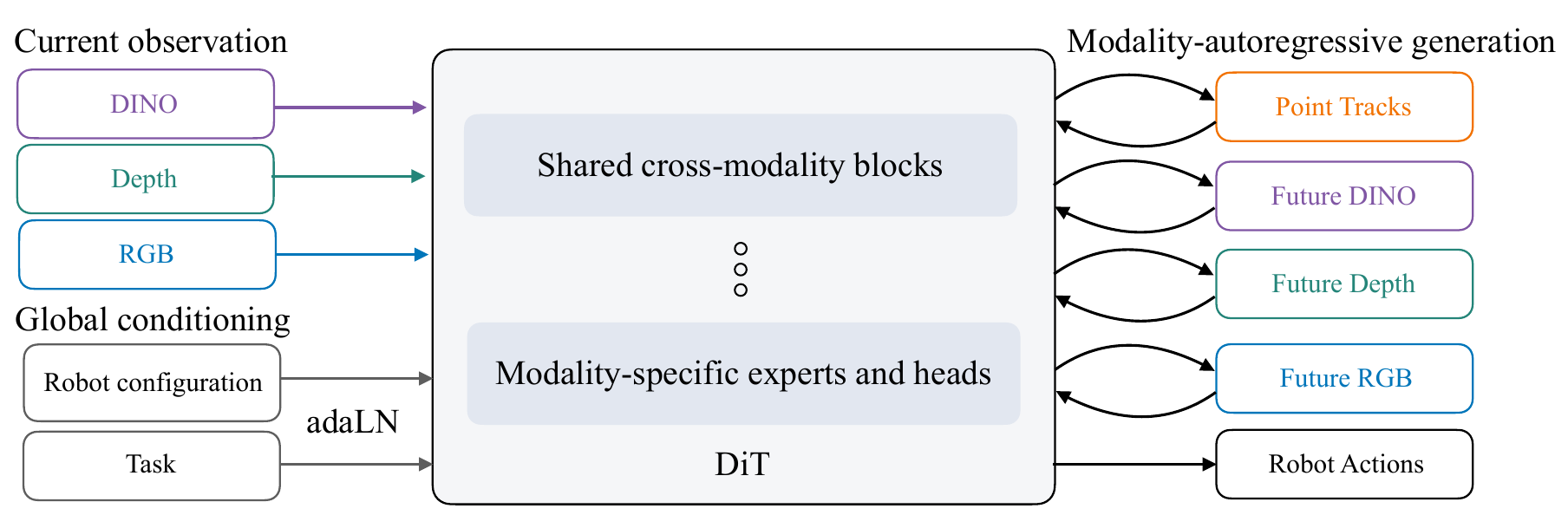}
    \caption{\textbf{\methodname\ architecture.}
    We embed the current observation from each modality and pass the resulting tokens into a shared diffusion transformer. The DiT combines cross-modal blocks with modality-specific experts and output heads. Robot configuration and a learned task embedding provide global adaLN conditioning to each DiT layer. \methodname{} sequentially denoises future modalities, conditioning each on earlier predictions, and denoises actions last.
    \vspace{-8pt}
    }
    \label{fig:modar_architecture}
\end{figure*}

\section{Introduction}

Future-observation prediction is a powerful objective for learning rich representations of the world's dynamics and semantics.
World-action models (WAMs) harness this objective for robotics by jointly modeling future observations and corresponding robot actions.

While robot action prediction requires \textit{action-labeled} robot data, future-observation prediction can learn from broader \textit{actionless} data sources or initialize from pretrained video generation models.
Supervision on these broader data sources can benefit action prediction both as a training-time auxiliary objective, even when future generation is omitted at deployment~\cite{yuan_fast-wam_2026}, and by enabling policies to condition action generation on predicted visual futures~\cite{du_learning_2023,li_turning_2026}.
Recent predictive world models have explored other representations of the future that emphasize physical, spatial, or semantic structure.
These include depth \cite{li_wam4d_2026}, point tracks \cite{guan_point_2026, hung_3pointr_2026, duisterhof_pointzero_2026, zhang_molmomotion_2026}, and pretrained visual features such as DINO \cite{zhou_dino-wm_2025, wang_st-wam_2026}.
Here, we use \textit{modality} to denote a representation of the future, 
including both sensory signals and derived features.
These modalities each contain unique inductive biases that capture 
manipulation-relevant features.
In this work, we ask: \textit{How should WAMs combine multiple modalities?}

Our contributions are as follows:
\textbf{(1)} We introduce \methodname, a WAM that autoregressively denoises
multiple future-observation modalities before generating actions.
\textbf{(2)} We contribute controlled experiments on how WAM formulation, target modalities, and actionless data scale affect policy performance. 
We find that modality-autoregressive generation performs best across all evaluated data scales and benefits most from additional actionless data. In our experiments, predicting point tracks, DINO features, and depth provide additive gains, while additionally predicting future RGB provides no consistent benefit.
We also fine-tune the 6B-parameter video-pretrained WAM Flex-$\pi$~\cite{yan_flex-_2026} on our data. Our 30.1M-parameter \methodname\ achieves a slightly higher observed average success rate (75\% vs.\ 72\%) despite using approximately
\(20\times\) fewer training FLOPs and no pretraining.
\textbf{(3)} We validate the resulting design on real-world bimanual manipulation using a mix of robot demonstrations and actionless human demonstrations.

%% file: 2_rw.tex
\begin{figure*}[t]
    \centering
    \includegraphics[width=\linewidth]{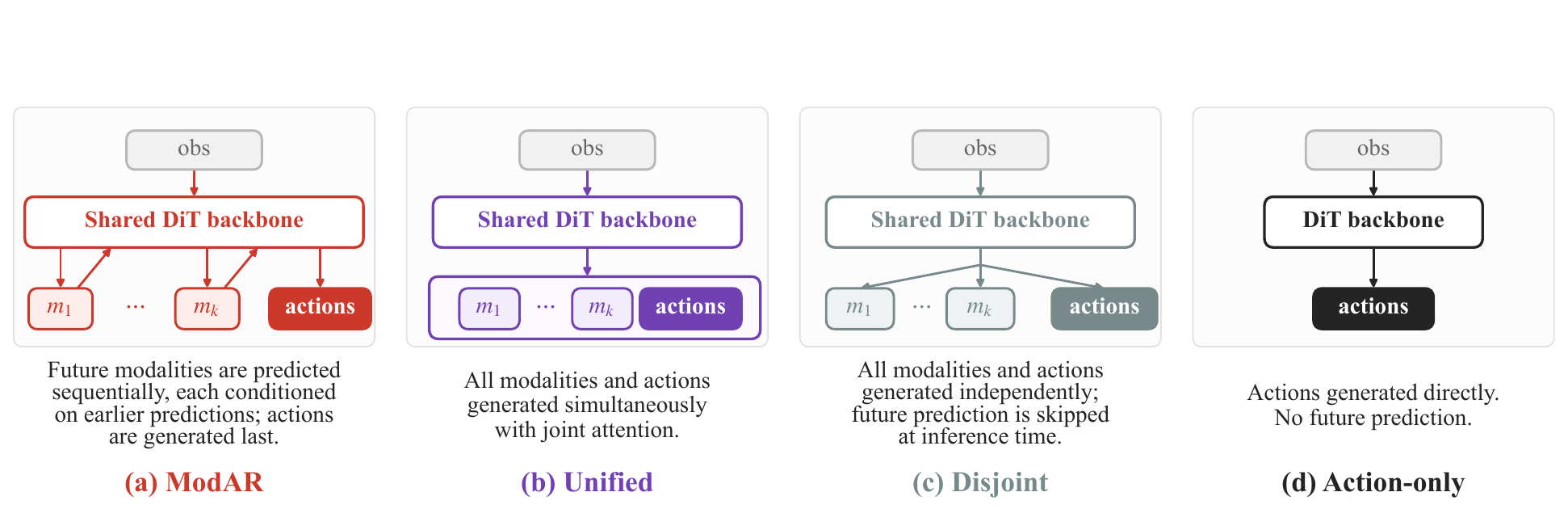}
    \vspace{-24pt}
    \caption{\textbf{WAM formulations.}
    We draw block diagrams of the evaluated WAM formulations and the Action-only baseline, which differ only in how they couple future prediction with action prediction.
    \methodname{} (a) autoregressively denoises future modalities before denoising actions.
    Unified (b) represents joint-generation WAMs such as DreamZero and Cosmos Policy~\cite{ye_world_2026,kim_cosmos_2026}, which denoise future-observation and action streams together.
    Disjoint (c) represents Fast-WAM~\cite{yuan_fast-wam_2026}, which predicts future observations and actions independently and omits future-observation prediction at inference time.
    Action-only (d) denoises actions directly without predicting future observations.
    Independent-noise (not drawn) is identical to Unified (b) except during training: rather than applying one shared noise level to all predicted streams, it samples each stream's flow timestep independently.
    \vspace{-4pt}
    }
    \label{fig:baselines}
\end{figure*}
\section{Related work}

\subsection{World-action models}

World-action models (WAMs) jointly model robot actions and future observations,
enabling large-scale training on both action-labeled robot demonstrations and
actionless demonstrations.
Their designs vary along two primary axes: how predicted futures inform action
generation and which representations of the future they predict.

WAM formulations differ in how predicted futures inform action generation
(Fig.~\ref{fig:baselines}).
Joint-generation models like DreamZero~\cite{ye_world_2026} co-denoise visual futures and
actions simultaneously with cross-stream attention (Fig.~\ref{fig:baselines}(b)).
Unified World Models~\cite{zhu_unified_2025} and Flex-$\pi$~\cite{yan_flex-_2026}
do the same, but
sample noise levels independently for different streams during training.
Fast-WAM~\cite{yuan_fast-wam_2026} instead prevents attention between future
and action targets (Fig.~\ref{fig:baselines}(c)), using future-observation prediction as a training-time auxiliary objective
and omitting it at deployment.
UniPi~\cite{du_learning_2023} and VERA~\cite{li_turning_2026} instead fully
denoise visual futures and then infer actions from the resulting frames,
rather than co-denoising both streams.
Building on this futures-then-actions ordering, \methodname{} autoregressively denoises multiple future modalities one at a time before denoising robot actions.
Our controlled experiments compare each formulation (Fig.~\ref{fig:baselines}) for multimodal WAMs and we find \methodname{} performs best. 

WAMs also differ in how they represent predicted futures.
Most WAMs predict future RGB as image latents from frozen video VAEs
\cite{zhu_unified_2025,pai_mimic-video_2025,kim_cosmos_2026,
ye_world_2026,li_causal_2026,yuan_fast-wam_2026}.
While this representation provides a convenient interface to pretrained video
generators, reconstructing visual appearance does not explicitly prioritize
the geometric, physical, and semantic structures most relevant to manipulation tasks.
Recent work has therefore explored predicting alternative modalities either
in place of RGB \cite{zhou_dino-wm_2025, li_egowam_2026} or alongside it
\cite{guan_point_2026, wang_st-wam_2026, li_wam4d_2026}.
Point tracks encode scene motion and correspondence independently of
appearance, providing direct supervision for how task-relevant scene elements
move through time
\cite{wen_any-point_2024, bharadhwaj_track2act_2024, hung_3pointr_2026}.
DINO features are robust to appearance variation while encoding object semantics
and scene structure~\cite{oquab_dinov2_2024}.
Depth makes scene geometry explicit and provides direct supervision for the
spatial reasoning required for predicting 3D actions~\cite{li_wam4d_2026}.
Concurrent work Flex-$\pi$ finds that jointly predicting multiple future modalities (RGB, 3D pointmaps, and DINO features) can improve performance over predicting future RGB \cite{yan_flex-_2026}.

Most existing WAMs build on pretrained video-generation models
\cite{kim_cosmos_2026, ye_world_2026, yan_flex-_2026}.
This initialization is highly effective, but makes it difficult to 
isolate the effects of WAM formulation, target representations, and actionless-data scale.
We instead train all models from scratch and systematically study these factors in a controlled setting.

\subsection{Multimodal generation}

Predicting multiple representations of the same underlying signal can provide
complementary supervision for representation learning.
By co-training a shared backbone across multiple objectives, each objective can enrich and regularize the representations used by the others.
Prior work finds that such co-training can improve per-task performance
relative to single-task training, and recent systems have scaled
this idea to many vision, language, and action tasks with strong results
\cite{bachmann_multimae_2022,mizrahi_4m_2023,lu_unified-io_2023,nvidia_cosmos_2026}.
Following this idea, we jointly predict multiple representations of the
future so that action prediction can benefit from their inductive biases.

Beyond the choice of targets, their generation order can influence how much
they help one another.
Latent Forcing~\cite{baade_latent_2026} generates image latents before pixels, allowing the latents to
serve as a semantic ``scratchpad'' for generating fine-grained appearance.
Similarly, Modality Forcing \cite{duisterhof_modality_2026} adapts a pretrained image generation model to jointly generate depth and finds that image models contain valuable
priors which improve depth accuracy.
These results suggest that \textit{generating easier-to-predict modalities earlier can expose
structure that simplifies subsequent predictions.}
We extend this principle to multimodal WAMs by generating modalities autoregressively,
starting with more structured modalities like point tracks and DINO features
followed by increasingly detailed depth and RGB predictions, and finally actions.

%% file: 3_method.tex
\begin{figure*}[t]
    \centering
    \begin{minipage}[t]{0.405\linewidth}
        \vspace{0pt}
        \centering
        \includegraphics[width=\linewidth,trim=8bp 0 5bp 0,clip]{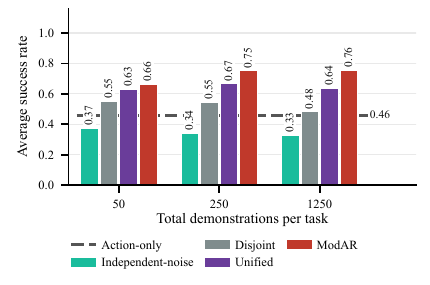}
        \vspace{-6mm}
        \par\smallskip{\small(a) Formulation comparison (Figure~\ref{fig:baselines})}
    \end{minipage}\hfill
    \begin{minipage}[t]{0.585\linewidth}
        \vspace{0pt}
        \centering
        \includegraphics[width=\linewidth,trim=24bp 0 8bp 0,clip]{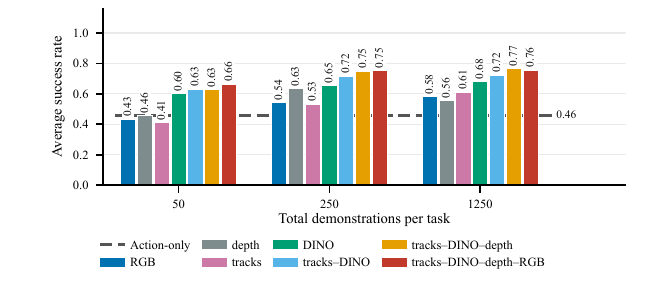}
        \vspace{-6mm}
        \par\smallskip{\small(b) Predicted modalities}
    \end{minipage}
    \caption{\textbf{RoboTwin simulation results.}
    Success rates over six tasks with 50 trials per task.
    We train all models with 50 action-labeled demonstrations per task, then add actionless demonstrations to increase the total demonstration count from 50 to 250 to 1,250.
    (a) \methodname\ achieves the highest average success rate among the
    formulations at each data scale.
    (b) We find it beneficial to predict future tracks, DINO, and depth, while additionally predicting future RGB provides no consistent gain.
    RGB (dark blue) matches the typical WAM setting of predicting only future RGB and actions.
    Independent-noise is identical to Unified except during training: rather than applying one shared noise level to all predicted streams, it samples each stream's flow timestep independently.
    \vspace{-8pt}
    }
    \label{fig:robotwin_simulation}
\end{figure*}

\section{Method}
\label{sec:method}

\subsection{Problem definition}

Given demonstrations spanning multiple manipulation tasks, our objective is to jointly model future
multimodal observations and robot actions conditioned on the current
observation, robot configuration, and task label.
We consider the set of predicted future modalities
\(\mathcal{M}=\{\mathrm{rgb},\mathrm{depth},\mathrm{dino},\mathrm{tracks}\}\).
At decision time \(t\), let \(\mathbf{o}_t\) denote the multimodal
observation.  The complete conditioning information is
\(\mathbf{c}_t=(\mathbf{o}_t,\,\mathbf{q}_t,\,g)\),
where \(\mathbf{q}_t\) is the robot's proprioceptive configuration (joint
positions and gripper opening) and \(g\) is the learned embedding associated
with a discrete task label. 
For each \(m\in\mathcal{M}\), let
\(\mathbf{Y}_t^m=(\mathbf{y}_{t+\Delta}^m,\ldots,
\mathbf{y}_{t+J\Delta}^m)\) denote \(J\) future targets in modality \(m\),
where \(\Delta\) is the dynamics stride and \(H=J\Delta\) is the prediction horizon.
We predict actions at every control step:
\(\mathbf{A}_t=(\mathbf{a}_t,\mathbf{a}_{t+1},\ldots,
\mathbf{a}_{t+H-1})\).  Thus, the final prediction target corresponds to the
next decision point \(t+H\), while \(\mathbf{A}_t\) contains the \(H\) actions
executed between \(t\) and \(t+H\).

\subsection{\methodname{}: Modality-Autoregressive World Modeling}

Given the targets above, \methodname{} denoises one future modality
at a time and uses each completed prediction as context for generating the next.
We generate actions last, conditioning
the policy on every generated modality.  
Formally, let \((m_1,\ldots,m_K)\) be an ordering of
\(\mathcal{M}\) (or a subset of \(\mathcal{M}\)), and write
\(\mathbf{Y}_t=(\mathbf{Y}_t^{m_1},\ldots,\mathbf{Y}_t^{m_K})\) for the
corresponding future targets.  We model their joint distribution with actions
as
\begin{equation}
p_\theta(\mathbf{Y}_t,\mathbf{A}_t \mid \mathbf{c}_t)
=p_\theta(\mathbf{A}_t\mid\mathbf{c}_t,\mathbf{Y}_t) \prod_{k=1}^{K}
p_\theta(\mathbf{Y}_t^{m_k}\mid\mathbf{c}_t,\mathbf{Y}_t^{<k})
,
\label{eq:modality_ar_factorization}
\end{equation}
where \(\mathbf{Y}_t^{<k}=(\mathbf{Y}_t^{m_1},\ldots,\mathbf{Y}_t^{m_{k-1}})\)
denotes the modalities preceding \(m_k\) in the generation order.
The final action-prediction step acts as an inverse dynamics model (IDM),
mapping the generated future observations to the actions that
induce the predicted transitions.
For actionless
examples, we omit action prediction and supervise only the available future
targets.

\textbf{Modality tokenization.}
We patchify RGB and depth maps.
We extract DINO tokens from the spatial patch tokens of a frozen DINOv2 encoder~\cite{oquab_dinov2_2024}.
For point tracks, we initialize a 2D grid of queries at patch centers and track them with CoTracker3~\cite{karaev_cotracker3_2024}; we represent each point-time pair with a token that encodes its displacement from its initial position and its visibility.
We linearly project each modality into the common token dimension.
We add a learned modality embedding and apply axial rotary position embeddings (RoPE)~\cite{su_roformer_2023} over time and the two spatial dimensions for visual tokens, and over time for action tokens.

\textbf{Shared world-action backbone.}
As shown in Fig.~\ref{fig:modar_architecture}, we process the resulting tokens
from all modalities with a diffusion transformer (DiT)~\cite{peebles_scalable_2023}.
The DiT first applies several shared transformer blocks that attend across all
causally available modality tokens.  It then applies a small modality-specific
expert stack, whose attention is restricted to that modality's stream, followed
by a linear output head.  The robot configuration, task embedding, and flow
timesteps for each modality condition the transformer through adaLN.  Thus,
cross-modal information is fused in the shared blocks before within-modality
specialization in the expert blocks.

\textbf{Block-causal modality generation.}
Equation~\eqref{eq:modality_ar_factorization} factorizes the joint multimodal distribution into a sequence of modality-conditional distributions. Our autoregressive sequence consists of modality blocks: we generate all tokens of one modality jointly before moving to the
next modality.  When generating block \(m_k\), the model attends
to the current observation and to the completed blocks
\(m_1,\ldots,m_{k-1}\); blocks later in the sequence are not available.  At
inference, we therefore denoise one block, re-embed the completed
prediction as context, and then denoise the next block.  
We use the order
tracks \(\rightarrow\) DINO \(\rightarrow\) depth
\(\rightarrow\) RGB. 
Intuitively, this orders the targets from compact, structured
representations that are easier to predict toward increasingly
high-dimensional and detailed representations.

During training, we supervise every stage in one pass by creating a clean
context copy and a noisy prediction copy of each target modality.  A
block-causal mask lets the prediction copy for \(m_k\) attend only to the
observation and context copies of \(m_1,\ldots,m_{k-1}\), preventing target
leakage.

\textbf{Injecting context noise.}
Modality-autoregressive generation is susceptible to cascading error, as imperfections in
earlier generations become inputs to all later predictions.  Following Latent
Forcing~\cite{baade_latent_2026}, we mitigate this by adding noise to context
blocks during training.  Specifically, when predicting \(m_k\), we
independently sample \(\boldsymbol{\epsilon}_j^{\mathrm{ctx}}\sim
\mathcal{N}(0,I)\) and
\(\tau_j^{\mathrm{ctx}}\sim\mathcal{U}(1-\beta,1)\) for every context block
\(m_j\) in every training example, and replace its clean target with
\(\widetilde{\mathbf{Y}}_{t,\mathrm{ctx}}^{m_j}
=\tau_j^{\mathrm{ctx}}\mathbf{Y}_t^{m_j}
+(1-\tau_j^{\mathrm{ctx}})\boldsymbol{\epsilon}_j^{\mathrm{ctx}}\).
We apply this noise only during training, not during inference.

\textbf{Training.}
We train every output stream with a JiT-style \(x\)-prediction objective~\cite{li_back_2026}.
Empirically, we find that replacing \(x\)-prediction with velocity (\(v\)) prediction is often unstable and can cause training to diverge.
Velocity prediction can struggle in high-dimensional spaces; predicting the
clean sample is well-conditioned when the data lie on a low-dimensional
manifold, as with images, depth, and other visual representations
\cite{li_back_2026,duisterhof_modality_2026}.
We use the same parameterization for robot action
generation~\cite{yang_abot-m0_2026,chen_mv-wam_2026}.
For each supervised modality \(m\), we independently sample
\(\boldsymbol{\epsilon}^m\sim\mathcal{N}(0,I)\) and a flow timestep \(\tau_m\), and form a linear flow-matching interpolant~\cite{lipman_flow_2023}
\begin{equation}
    \widetilde{\mathbf{Y}}_t^m
    = \tau_m\mathbf{Y}_t^m + (1-\tau_m)\boldsymbol{\epsilon}^m .
\end{equation}
The corresponding output head directly predicts the clean target
\(\widehat{\mathbf{Y}}_t^m\). We minimize the loss function:  
\begin{equation}
    \mathcal{L}_m(\theta)
    =
    \mathbb{E}\!\left[
    \frac{\lVert\widehat{\mathbf{Y}}_t^m-\mathbf{Y}_t^m\rVert_2^2}
    {\max(1-\tau_m,\delta_m)^2}
    \right],
\end{equation}
where \(\delta_m\) stabilizes the loss near the clean endpoint.  We apply the
same objective to the action chunk, replacing
\((\mathbf{Y}_t^m,\widehat{\mathbf{Y}}_t^m,\tau_m,\delta_m)\) with
\((\mathbf{A}_t,\widehat{\mathbf{A}}_t,\tau_{\mathrm{act}},
\delta_{\mathrm{act}})\).
The total loss is
\begin{equation}
    \mathcal{L}(\theta)
    = \sum_{k=1}^{K}
      \mathcal{L}_{m_k}
      + \mathcal{L}_{\mathrm{act}},
    \label{eq:modar_loss}
\end{equation}
We include the action term only for examples with action labels.

\textbf{Inference.}
We generate one modality at a time in the order \(m_1,\ldots,m_K\). We initialize
each stream from isotropic Gaussian noise and integrate it from \(\tau=0\) to \(1\) using
the velocity implied by its clean prediction,
\(\mathbf{v}_\theta=(\widehat{\mathbf{Y}}-\widetilde{\mathbf{Y}})/(1-\tau)\);
the same equation applies to the action stream with \(\mathbf{Y}\) replaced by
\(\mathbf{A}\).
The resulting clean tokens become context for generating the
next modality, and we condition the final action chunk on the complete
generated future. We cache keys and values for the current observation and
all previously generated modalities to avoid recomputing them at every denoising step.

\input{tables/pertask}

%% file: tables/pertask.tex
\begin{table*}[t]
\centering
\input{tables/arch_pertask}\hfill
\input{tables/mod_pertask}
\end{table*}

%% file: tables/arch_pertask.tex
\begin{minipage}[t]{0.43\textwidth}
\vspace{0pt}
\centering
\captionof{table}{Per-task success rates by WAM formulation at
total-demonstration scale \(D\): action-only, independent-noise, disjoint,
unified, and \methodname.}
\label{tab:arch_pertask}
\fontsize{5.8}{5.9}\selectfont
\renewcommand{\arraystretch}{0.82}
\setlength{\tabcolsep}{1.5pt}
\begin{tabular}{@{}llccccc@{}}
\toprule
Task & \(D\) & \shortstack{Action-\\only} &
\shortstack{Independent-\\noise} & Disjoint & Unified & \methodname \\
\midrule
\multirow{3}{*}{dump bin}
  & 50   & 0.82 & 0.64 & \second{0.90} & 0.88 & \best{0.92} \\
  & 250  & \textemdash & 0.68 & \second{0.92} & \second{0.92} & \best{0.94} \\
  & 1250 & \textemdash & 0.64 & 0.78 & \second{0.94} & \best{0.98} \\
\cmidrule(lr){1-7}
\multirow{3}{*}{pick bottles}
  & 50   & 0.30 & 0.40 & 0.44 & \best{0.58} & \second{0.48} \\
  & 250  & \textemdash & 0.38 & \second{0.42} & \best{0.60} & \best{0.60} \\
  & 1250 & \textemdash & 0.38 & 0.38 & \best{0.62} & \second{0.58} \\
\cmidrule(lr){1-7}
\multirow{3}{*}{place bread}
  & 50   & 0.24 & 0.20 & \second{0.46} & 0.44 & \best{0.52} \\
  & 250  & \textemdash & 0.14 & 0.48 & \second{0.50} & \best{0.74} \\
  & 1250 & \textemdash & 0.28 & 0.22 & \second{0.44} & \best{0.72} \\
\cmidrule(lr){1-7}
\multirow{3}{*}{put bottles}
  & 50   & 0.32 & 0.20 & 0.36 & \second{0.52} & \best{0.64} \\
  & 250  & \textemdash & 0.20 & 0.36 & \second{0.62} & \best{0.82} \\
  & 1250 & \textemdash & 0.18 & 0.42 & \second{0.56} & \best{0.82} \\
\cmidrule(lr){1-7}
\multirow{3}{*}{stack bowls}
  & 50   & \second{0.64} & 0.36 & 0.58 & \best{0.76} & \best{0.76} \\
  & 250  & \textemdash & 0.18 & 0.62 & \best{0.76} & \second{0.68} \\
  & 1250 & \textemdash & 0.04 & 0.62 & \second{0.68} & \best{0.80} \\
\cmidrule(lr){1-7}
\multirow{3}{*}{turn switch}
  & 50   & 0.44 & 0.44 & 0.58 & \second{0.60} & \best{0.66} \\
  & 250  & \textemdash & 0.48 & 0.48 & \second{0.60} & \best{0.74} \\
  & 1250 & \textemdash & 0.46 & 0.48 & \second{0.58} & \best{0.64} \\
\midrule
\multirow{3}{*}{overall}
  & 50   & 0.46 & 0.37 & 0.55 & \second{0.63} & \best{0.66} \\
  & 250  & \textemdash & 0.34 & 0.55 & \second{0.67} & \best{0.75} \\
  & 1250 & \textemdash & 0.33 & 0.48 & \second{0.64} & \best{0.76} \\
\bottomrule
\end{tabular}
\end{minipage}%

%% file: tables/mod_pertask.tex
\begin{minipage}[t]{0.56\textwidth}
\vspace{0pt}
\centering
\captionof{table}{Per-task success rates by predicted-modality subset at
total-demonstration scale \(D\). Modalities comprise RGB, depth, point tracks,
and DINO features.}
\label{tab:mod_pertask}
\fontsize{5.8}{5.9}\selectfont
\renewcommand{\arraystretch}{0.82}
\setlength{\tabcolsep}{1.5pt}
\begin{tabular}{@{}llcccccccc@{}}
\toprule
Task & \(D\) & \shortstack{Action-\\only} & RGB & Depth & Tracks & DINO &
\shortstack{Tracks\\+ DINO} & \shortstack{Tracks + DINO\\+ Depth} &
\shortstack{Tracks + DINO\\+ Depth + RGB} \\
\midrule
\multirow{3}{*}{dump bin}
  & 50   & 0.82 & 0.80 & 0.90 & 0.90 & \best{0.96} & \second{0.92} & \second{0.92} & \second{0.92} \\
  & 250  & \textemdash & \second{0.94} & 0.92 & 0.92 & \second{0.94} & \second{0.94} & \best{0.96} & \second{0.94} \\
  & 1250 & \textemdash & \second{0.98} & 0.88 & 0.90 & 0.96 & \best{1.00} & 0.94 & \second{0.98} \\
\cmidrule(lr){1-10}
\multirow{3}{*}{pick bottles}
  & 50   & 0.30 & 0.18 & 0.44 & 0.12 & 0.44 & \best{0.58} & \second{0.48} & \second{0.48} \\
  & 250  & \textemdash & 0.32 & 0.48 & 0.48 & \best{0.62} & \best{0.62} & 0.58 & \second{0.60} \\
  & 1250 & \textemdash & 0.30 & 0.40 & 0.42 & 0.62 & \best{0.70} & \second{0.66} & 0.58 \\
\cmidrule(lr){1-10}
\multirow{3}{*}{place bread}
  & 50   & 0.24 & 0.26 & 0.38 & 0.10 & 0.44 & \second{0.52} & \best{0.54} & \second{0.52} \\
  & 250  & \textemdash & 0.34 & 0.64 & 0.26 & 0.56 & 0.64 & \best{0.76} & \second{0.74} \\
  & 1250 & \textemdash & 0.36 & 0.60 & 0.38 & 0.54 & \second{0.64} & \best{0.72} & \best{0.72} \\
\cmidrule(lr){1-10}
\multirow{3}{*}{put bottles}
  & 50   & 0.32 & 0.20 & 0.12 & 0.30 & \best{0.64} & \second{0.48} & \second{0.48} & \best{0.64} \\
  & 250  & \textemdash & 0.38 & 0.48 & 0.48 & 0.60 & 0.66 & \second{0.74} & \best{0.82} \\
  & 1250 & \textemdash & 0.54 & 0.42 & 0.74 & 0.64 & 0.66 & \best{0.90} & \second{0.82} \\
\cmidrule(lr){1-10}
\multirow{3}{*}{stack bowls}
  & 50   & 0.64 & 0.58 & 0.54 & \second{0.74} & 0.66 & 0.70 & 0.72 & \best{0.76} \\
  & 250  & \textemdash & 0.72 & 0.74 & 0.64 & 0.66 & \best{0.86} & \second{0.78} & 0.68 \\
  & 1250 & \textemdash & 0.66 & 0.64 & 0.66 & \best{0.84} & 0.74 & \second{0.80} & \second{0.80} \\
\cmidrule(lr){1-10}
\multirow{3}{*}{turn switch}
  & 50   & 0.44 & 0.58 & 0.36 & 0.30 & 0.46 & 0.58 & \second{0.62} & \best{0.66} \\
  & 250  & \textemdash & 0.56 & 0.54 & 0.42 & 0.54 & 0.58 & \second{0.68} & \best{0.74} \\
  & 1250 & \textemdash & \best{0.64} & 0.40 & 0.56 & 0.48 & \second{0.60} & 0.58 & \best{0.64} \\
\midrule
\multirow{3}{*}{overall}
  & 50   & 0.46 & 0.43 & 0.46 & 0.41 & 0.60 & \second{0.63} & \second{0.63} & \best{0.66} \\
  & 250  & \textemdash & 0.54 & 0.63 & 0.53 & 0.65 & \second{0.72} & \best{0.75} & \best{0.75} \\
  & 1250 & \textemdash & 0.58 & 0.56 & 0.61 & 0.68 & 0.72 & \best{0.77} & \second{0.76} \\
\bottomrule
\end{tabular}
\end{minipage}

%% file: 4_experiments.tex
\section{Experiments}

In simulation, we evaluate WAM formulations (Figure~\ref{fig:baselines}), actionless-data scaling, and predicted
modality sets. We also compare with a
large video-model-initialized WAM, Flex-$\pi$~\cite{yan_flex-_2026}. 
We then evaluate \methodname\ on real-world bimanual
manipulation and learning from human demonstrations.

\subsection{Experimental setup}

\textbf{Simulation.}
We evaluate on six representative RoboTwin~\cite{chen_robotwin_2025} tasks: ``dump bin'', ``pick bottles'',
``place bread'', ``put bottles'', ``stack bowls'', and ``turn switch.'' 
Each task uses
\(D\in\{50,250,1250\}\) total training demonstrations. We retain 50 action-labeled demonstrations
and use the remaining \(D-50\) as actionless demonstrations.
For each method and data scale, we train one multitask model for all
six tasks.
We evaluate all models on 50 held-out initial
conditions per task.

\textbf{Real world.}
We evaluate on three challenging tabletop tasks---stacking cups, folding a crumpled
towel, and placing an object in a drawer and closing the drawer---using bimanual YAM arms
(Fig.~\ref{fig:teaser}).
For each task, we collect 100 teleoperated robot demonstrations, 200
in-domain actionless human demonstrations, and 1{,}000 EgoDex demonstrations~\cite{hoque_egodex_2025}.
We use the corresponding EgoDex categories ``stack/unstack cups,'' ``basic
fold,'' and ``insert/remove drawer'' for cup stacking, towel folding, and
drawer placement, respectively. 
For each method and data mixture, we train one multitask model for
all three real-world tasks.
We evaluate each reported method and data mixture with 30 rollouts per task,
with varied initial object poses.

\begin{figure*}[t]
    \begin{minipage}[t]{0.49\linewidth}
        \centering
        \includegraphics[width=\linewidth]{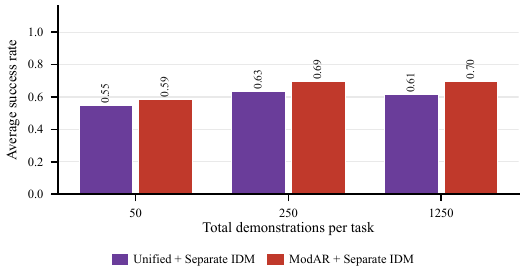}
        \captionof{figure}{
        \textbf{Evaluating predicted futures with a separate IDM.}
        Recall from Sec.~\ref{sec:method} that \methodname's final
        action-prediction step acts as an
        inverse-dynamics model (IDM), mapping its future-observation predictions
        to actions.  In contrast, Unified denoises futures and actions jointly,
        so its action prediction conditions on partially noisy predicted
        futures.
        To remove this asymmetry, we discard both models' native action predictions and
        pass their predicted futures to the same separately
        trained IDM.  \methodname\ still outperforms
        Unified, showing that its predicted futures are themselves more useful
        for predicting actions.
        }
        \label{fig:shared_idm}
    \end{minipage}\hfill
    \begin{minipage}[t]{0.49\linewidth}
        \centering
        \includegraphics[width=\linewidth]{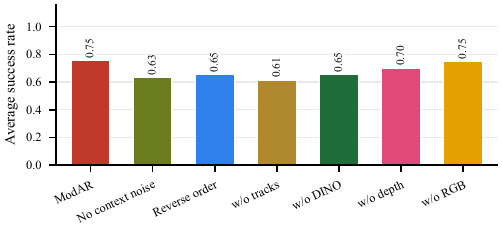}
        \captionof{figure}{
        \textbf{\methodname\ ablations in RoboTwin.}
        Average success rate over six tasks, with 250 total
        demonstrations per task and 50 action-labeled demonstrations.
        We ablate context noise (\textit{No context noise}),
        generation order (\textit{Reverse order}), and each individual
        future-observation modality.
        Results suggest that context noise is critical for robust autoregressive generation.
        Generating modalities in reverse order also reduces performance, supporting our hypothesis that more structured modalities act as ``scratchpads'' for more detailed modalities.
        Removing future RGB prediction does not reduce performance, suggesting that RGB contributes the least of the four future-observation modalities.
        }
        \label{fig:ablations}
    \end{minipage}
    \vspace{-8pt}
\end{figure*}

\setcounter{dbltopnumber}{1}
\begin{figure*}[t]
    \centering
    \begin{minipage}[t]{0.49\linewidth}
        \vspace{0pt}
        \centering
        \includegraphics[width=\linewidth]{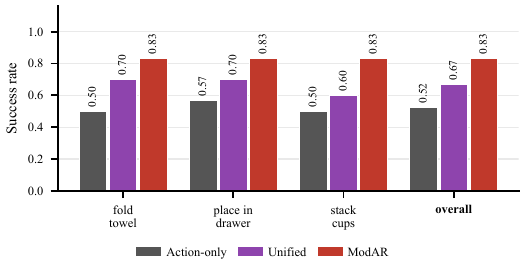}
        \par\smallskip\textbf{(a) Formulation}
    \end{minipage}\hfill
    \begin{minipage}[t]{0.49\linewidth}
        \vspace{0pt}
        \centering
        \includegraphics[width=\linewidth]{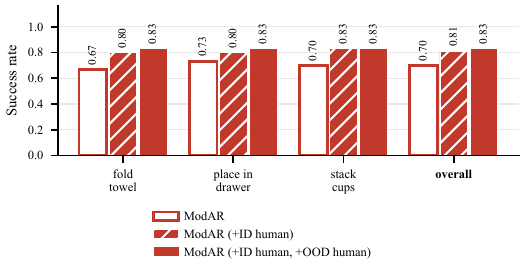}
        \par\smallskip\textbf{(b) Actionless human data}
    \end{minipage}
    \caption{\textbf{Real-world results.}
    Success rates on real-world tasks, with 30 trials per model per task.
    (a) All methods use 100 robot demonstrations per task; \methodname\ and
    Unified additionally use 200 in-domain actionless human demonstrations and
    1{,}000 out-of-domain EgoDex demonstrations.  \methodname\ achieves the
    highest success rate on all three tasks.
    (b) We train separate \methodname\ models with just 100 robot
    demonstrations per task, then adding 200 in-domain actionless human
    demonstrations (used to supervise future-observation predictions, but not
    action prediction), and then also adding 1{,}000 out-of-domain human
    demonstrations from EgoDex (used similarly). The average success rate
    progressively improves from 70.0\% to 81.1\% to 83.3\%.
    \vspace{-8pt}
    }
    \label{fig:real_world}
\end{figure*}

\textbf{Baselines.}
We compare representative WAM formulations (Fig.~\ref{fig:baselines}) within one controlled implementation.
All methods share the same backbone, action-labeled data, and
optimization budget.

\noindent\textit{Unified} (Fig.~\ref{fig:baselines}(b)) jointly denoises all future-observation and action streams using
one shared flow timestep, representing joint-generation WAMs
such as DreamZero and Cosmos Policy~\cite{ye_world_2026,kim_cosmos_2026}.

\noindent\textit{Disjoint} (Fig.~\ref{fig:baselines}(c)) predicts each target independently,
without attention between future-observation and action targets, as in Fast-WAM~\cite{yuan_fast-wam_2026}.

\noindent\textit{Independent-noise} samples a
separate flow timestep for every stream during training, as in Unified World Models and
Flex-$\pi$~\cite{zhu_unified_2025,yan_flex-_2026}; at inference, it uses the same simultaneous
denoising process as Unified (Fig.~\ref{fig:baselines}(b)).

\noindent\textit{Action-only} (Fig.~\ref{fig:baselines}(d)) predicts actions directly from
observations without future-observation prediction,
representing a typical flow-matching behavior cloning policy. 
Because it has
no future-observation prediction objective, Action-only trains only on action-labeled
data.

\subsection{Simulation results}

Figure~\ref{fig:robotwin_simulation}
compares WAM formulations and predicted-modality subsets as the
amount of actionless data varies.

\textbf{Formulation comparison.}
Figure~\ref{fig:robotwin_simulation}(a) compares formulations trained to
predict all four modalities; Table~\ref{tab:arch_pertask} gives the per-task
breakdown.  \methodname\ achieves the highest average success rate at every data scale.
We hypothesize that early modalities act as ``scratchpads'' for later ones:
generating coarser or easier targets first provides structured context for
more detailed targets \cite{baade_latent_2026}.

Independent-noise generation performs poorly in our from-scratch experiments.
Because independently sampled noise levels rarely match the synchronized
test-time denoising schedule, we hypothesize that the model receives
insufficient training signal near its inference regime.

\textbf{Scaling with actionless data.}
\methodname\ benefits most from additional actionless data: adding 1,200
actionless demonstrations improves the average success rate from 66\% to 76\%
(10 percentage points), compared with a mere 1\% improvement for Unified.
Disjoint generation outperforms Action-only prediction on average, but its performance decreases as we add actionless data.  One
possible explanation is that with more actionless data, the shared representation becomes increasingly
shaped by the future-observation prediction objective, causing negative transfer to the action prediction objective. 

\textbf{Modality comparison.}
Figure~\ref{fig:robotwin_simulation}(b) compares \methodname\ variants that
predict each target modality individually against variants that predict progressively
larger modality sets, following the generation order point tracks, DINO,
depth, and RGB; Table~\ref{tab:mod_pertask} gives the per-task breakdown.
Across data scales, performance generally holds or increases with each added
modality, showing that \methodname\ can effectively combine the benefits of predicting multiple modalities.
However, the best modality set naturally varies across tasks because different features define each task, and some modalities represent those features better than others; as a result, additional modalities do not always help.
In particular, additionally predicting future RGB on top of the first three modalities provides
no consistent gain.
We hypothesize that predicting future RGB introduces
high-variance appearance details while adding little information beyond the
more structured targets in our setting.

\textbf{Separate inverse-dynamics model.}
Recall from Sec.~\ref{sec:method} that \methodname's final action-prediction
step acts as an inverse-dynamics model (IDM), mapping its fully generated
future-observation predictions to actions.  Unified instead predicts actions
while its predicted futures are still being jointly denoised.  Thus,
\methodname\ could outperform Unified simply because its action predictor
conditions on more informative predicted futures.  To test this, we evaluate both models using the same
separately trained IDM in place of their native action predictors.
We train this IDM
to map ground-truth future observations to actions, and apply context noise to its inputs during training.
We evaluate the best-performing \methodname\ and Unified checkpoints on 50 held-out initial conditions per task using this separate IDM.
At each prediction step, each model generates its future-observation predictions and
actions normally.  We then discard its native action prediction, pass its predicted
futures to the separate IDM, and execute the resulting actions
(Fig.~\ref{fig:shared_idm}).
\methodname\ still outperforms Unified with the separate IDM, suggesting that
its predicted futures are themselves more useful for predicting actions.

\textbf{Sampling-step comparison.}
\methodname\ generates each modality autoregressively, and performs 8 Euler
integration steps per modality---40 steps across four future-observation streams and actions---whereas Unified, Independent-noise, and Disjoint use only 8 steps in total.
To test whether \methodname's gains arise simply from its larger sampling budget, we re-evaluate each formulation at \(D=250\) with 40 Euler steps, matching \methodname's total number of sampling steps.
Additional steps do not close the gap: Unified decreases from 67\% to 59\%, Disjoint from 55\% to 54\%, and Independent-noise increases from 34\% to 37\%, compared to \methodname's 75\%.

\textbf{Comparison to a video-model-initialized WAM.}
In addition to our controlled from-scratch experiments, we compare against concurrent work Flex-$\pi$~\cite{yan_flex-_2026} at \(D=250\).
We adapt the official implementation to our single head-camera setting and to an action horizon of \(H=16\), with visual targets at \(t+\{4,8,12,16\}\).
Following their recipe, we initialize the video backbone from Wan2.2-TI2V-5B~\cite{wan_wan_2025,wan_wan22_ti2v_2025} and interpolate those weights to a smaller hidden size for the action expert, then full-fine-tune all trainable Flex-$\pi$ components while keeping the VAE, text encoder, and DINOv3 encoder~\cite{simeoni_dinov3_2026} frozen.
On actionless demonstrations we retain every visual objective (RGB latents, pointmaps, and DINO features) and mask the action loss.

We train with global batch size 288 and evaluate on the same 50 held-out initial conditions per task used in our other experiments.
At inference we jointly denoise all visual streams and actions, i.e., Flex-$\pi$'s full multi-stream generation mode.
At its final checkpoint after 30{,}000 steps, Flex-$\pi$ attains an average success rate of \textbf{72\%}.
Compared to our trained-from-scratch tracks--DINO--depth \methodname\ variant, which achieves 75\% average success,
Flex-$\pi$ has approximately \(200\times\) as many parameters (6B vs 30.1M) and uses approximately \(20\times\) as many training FLOPs.
This comparison shows that, on these in-distribution tasks, a compact WAM
trained from scratch can perform as well or better than a much larger video-model-initialized WAM using substantially less training compute.
We note that this is a system-level comparison rather than a controlled architectural comparison, since the models differ in scale, pretraining, target modalities, and training recipe.

\textbf{Inference latency.}
On a single NVIDIA GeForce RTX 5090 GPU, end-to-end inference for \methodname{} takes \(147.9\) ms (\(6.76\) Hz) when generating all four future-observation modalities and actions.

\subsection{Ablations}
\label{sec:ablation}
Figure~\ref{fig:ablations} ablates the \methodname\ design at the 250-demonstration
scale, where the full model achieves a 75\% average success rate.
\textbf{Context noise.}
First, we remove
the context noise added to previously generated modalities during
training (\textit{No context noise}).
The success rate falls to 63\%, showing that context noise during training is critical for preventing errors from compounding across successive modalities.

\textbf{Modality order.}
We also reverse the future-modality order to RGB \(\rightarrow\) depth
\(\rightarrow\) DINO \(\rightarrow\) tracks (actions are still generated last).  This lowers the average success
rate to 65\%, supporting our choice to generate compact, structured
representations before increasingly detailed ones. 
We leave a full systematic comparison of modality orderings to future work.

\textbf{Ablating modalities.}
Finally, we measure the contribution of each future-observation modality by
removing one modality at a time from the full model.
Removing tracks (\textit{w/o tracks}), DINO (\textit{w/o DINO}), or
depth (\textit{w/o depth})
lowers the average success rate to 61\%, 65\%, and 70\%, respectively,
whereas removing future RGB prediction leaves it unchanged.
Thus,
predicting RGB contributes the least of the four modalities to success rate in this setting.

\subsection{Real-world experiments}

Because adding RGB provided no consistent benefit in simulation while
increasing training and generation costs, we train all real-world WAMs to observe and predict
only tracks, DINO, and depth.

\textbf{Formulation comparison.}
Results in Fig.~\ref{fig:real_world}(a) mirror the simulation results:
across 30 trials per task, \methodname\ achieves the highest success rate on all
three tasks and an overall success rate of 83.3\%, compared with 66.7\% for
Unified and 52.2\% for Action-only.

\textbf{Effect of adding human demonstration data.}
We compare \methodname\ trained on only the 100 teleoperated robot demonstrations per
task against adding 200 in-domain human demonstrations and then also adding
1{,}000 out-of-domain EgoDex demonstrations.
These additions improve the average success rate from 70.0\% to 81.1\% and
83.3\%, respectively (Fig.~\ref{fig:real_world}(b)),
indicating that \methodname\ can benefit from actionless data collected
across embodiments and domains.

\subsection{Implementation details}

\textbf{Inputs and outputs.}
Models condition on a single-camera \(168\times224\) observation and predict future
observations with horizon \(H=16\) and dynamics stride \(\Delta=8\), yielding
\(J=2\) sparse targets at \(t+8\) and \(t+16\), together with a dense
\(H\)-step action chunk; policies replan every \(H\) steps.
Actions and robot configurations are 14-dimensional absolute dual-arm joint
configurations. RGB, depth, and track queries use a \(12\times16\) grid of
\(14\times14\) patches. We obtain point tracks by tracking the patch-center
queries with CoTracker3~\cite{karaev_cotracker3_2024}, and extract DINO targets
with a frozen DINOv2 ViT-S/14 encoder~\cite{oquab_dinov2_2024}.

\textbf{Architecture.}
The shared DiT contains six
width-384 transformer blocks with six attention heads (ViT-S), followed by
two width-384 expert blocks for DINO, depth, and RGB, one width-128 block for
tracks, and two width-128 blocks for actions.

\textbf{Optimization.}
We use AdamW with learning
rate \(10^{-4}\), betas \((0.9,0.95)\), weight decay \(0.1\), gradient
clipping at \(1.0\), and global batch size 48. Training uses bfloat16, an
EMA with decay \(0.999\), and a 48{,}000-sample linear warmup followed by a
constant learning rate. We train all methods for 1.2M optimizer steps.
At each optimizer step, we sample equal-sized batches from the action-labeled
and actionless pools and weight
their losses equally.
We evaluate checkpoints every 100{,}000 optimizer steps and report the
best-checkpoint success rate on 50 held-out initial conditions per task.

\textbf{Flow and sampling.}
For \methodname, we set the maximum context-noise level to \(\beta=0.5\),
set \(\delta_m=\delta_{\mathrm{act}}=0.05\), use unit loss weights, and sample
logit-normal flow timesteps with
\((\mu,\sigma)=(-2,1),(0,1),(-1,1.6),(-1,1),(-1,1)\) for DINO, tracks,
depth, RGB, and actions, respectively. Unified uses one shared flow timestep
with \((\mu,\sigma)=(-1,1)\); Independent-noise samples each stream's
modality-specific distribution independently. We integrate each generated
stream with eight Euler steps.

\textbf{Real-world data collection.}
We collect robot demonstrations by teleoperation with paired teacher arms
using the RAIDEN toolkit~\cite{iwase_raiden_2026}. A fixed third-person ZED
stereo camera provides RGB observations and stereo depth for both robot and
actionless human demonstrations.

%% file: 5_conclusion.tex
\section{Conclusion}

We introduce \methodname, a world-action model that autoregressively denoises
multiple future-observation modalities before predicting actions.  In a controlled
simulation study, this formulation achieves the highest average success rate among representative existing WAM formulations and improves the most as we add actionless data; on three real-world tasks, it achieves the highest success rate among baselines and also improves with human video demonstrations.
On average, point tracks, DINO features, and depth provide complementary gains,
whereas additionally predicting future RGB provides no consistent benefit in our experiments.
These findings highlight modality-autoregressive prediction as a
promising alternative to relying solely on future RGB prediction for world-action modeling.

\section{Limitations}

Our experiments cover a limited number of tasks and use discrete task labels
rather than language instructions, so they do not establish broad
generalization across tasks, objects, or scenes.
Sequentially generating multiple modalities also increases
inference latency relative to simultaneous or action-only generation.
Future work should explore larger-scale experiments with broader task diversity, generalization evaluation, and training on heterogeneous internet-scale actionless data.
Future work can also explore the best order for generating modalities, which may depend on the task or specific scenario.

\section{Acknowledgment}
We thank Narek Harutyunyan for assistance in data collection.

This material is based upon work supported by the National Science Foundation Graduate Research Fellowship Program under Grant Nos. DGE2140739 and DGE2631988. Any opinions, findings, and conclusions or recommendations expressed in this material are those of the author(s) and do not necessarily reflect the views of the National Science Foundation.

This work used Bridges-2 at Pittsburgh Supercomputing Center through allocation CIS260202p from the Advanced Cyberinfrastructure Coordination Ecosystem: Services \& Support (ACCESS) program, which is supported by National Science Foundation grants \#2138259, \#2138286, \#2138307, \#2137603, and \#2138296~\cite{brown_bridges2_2021,boerner_access_2023}.